\documentclass[sigconf]{acmart}

\AtBeginDocument{%
  \providecommand\BibTeX{{%
    \normalfont B\kern-0.5em{\scshape i\kern-0.25em b}\kern-0.8em\TeX}}}

\usepackage{graphicx}
\usepackage{amsmath}
\usepackage{booktabs}
\usepackage{tabularray}
\usepackage{pifont}
\usepackage{diagbox}
\usepackage{adjustbox}
\usepackage{multirow}
\usepackage{cancel}
\usepackage{balance}
\usepackage{appendix}

\newcommand{\xmark}{\ding{56}}
\newcommand{\cmark}{\ding{51}}

\copyrightyear{2024} 
\acmYear{2024} 
\setcopyright{rightsretained} 
\acmConference[KDD '24]{Proceedings of the 30th ACM SIGKDD Conference on Knowledge Discovery and Data Mining}{August 25--29, 2024}{Barcelona, Spain}
\acmBooktitle{Proceedings of the 30th ACM SIGKDD Conference on Knowledge Discovery and Data Mining (KDD '24), August 25--29, 2024, Barcelona, Spain}\acmDOI{10.1145/3637528.3672004}
\acmISBN{979-8-4007-0490-1/24/08}

\makeatletter
\gdef\@copyrightpermission{
 \begin{minipage}{0.3\columnwidth}
  \href{https://creativecommons.org/licenses/by/4.0/}{\includegraphics[width=0.90\textwidth]{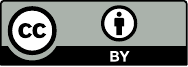}}
 \end{minipage}\hfill
 \begin{minipage}{0.7\columnwidth}
  \href{https://creativecommons.org/licenses/by/4.0/}{This work is licensed under a Creative Commons Attribution International 4.0 License.}
 \end{minipage}
 \vspace{5pt}
}
\makeatother

\begin{document}

\title{Image Similarity Using an Ensemble of Context-Sensitive Models}


\author{Zukang Liao}
\affiliation{%
  \institution{University of Oxford}
  \city{Oxford}
  \country{United Kingdom}}
\email{zukang.liao@eng.ox.ac.uk}

\author{Min Chen}
\affiliation{%
  \institution{University of Oxford}
  \city{Oxford}
  \country{United Kingdom}}
\email{min.chen@oerc.ox.ac.uk}


\begin{abstract}
Image similarity has been extensively studied in computer vision. In recent years, machine-learned models have shown their ability to encode more semantics than traditional multivariate metrics. However, in labelling semantic similarity, assigning a numerical score to a pair of images is impractical, making the improvement and comparisons on the task difficult. In this work, we present a more intuitive approach to build and compare image similarity models based on labelled data in the form of A:R vs B:R, i.e., determining if an image A is closer to a reference image R than another image B. We address the challenges of sparse sampling in the image space (R, A, B) and biases in the models trained with context-based data by using an ensemble model.
Our testing results show that the ensemble model constructed performs $\sim5\%$ better than the best individual context-sensitive models. They also performed better than the models that were directly fine-tuned using mixed imagery data as well as existing deep embeddings, e.g., CLIP~\cite{clip} and DINO~\cite{dino}. This work demonstrates that context-based labelling and model training can be effective when an appropriate ensemble approach is used to alleviate the limitation due to sparse sampling.

\end{abstract}

\ccsdesc[500]{Computing methodologies~Image representations}
\ccsdesc[300]{Computing methodologies~Ranking}
\ccsdesc[100]{Computing methodologies~Cross-validation}

\keywords{Image Similarity, Context-Sensitive Models, Analytical Ensemble Method, Semantic Distance, Global Features, Dataset}


\maketitle


\begin{figure}[t]
    \centering
    \includegraphics[height=45mm]{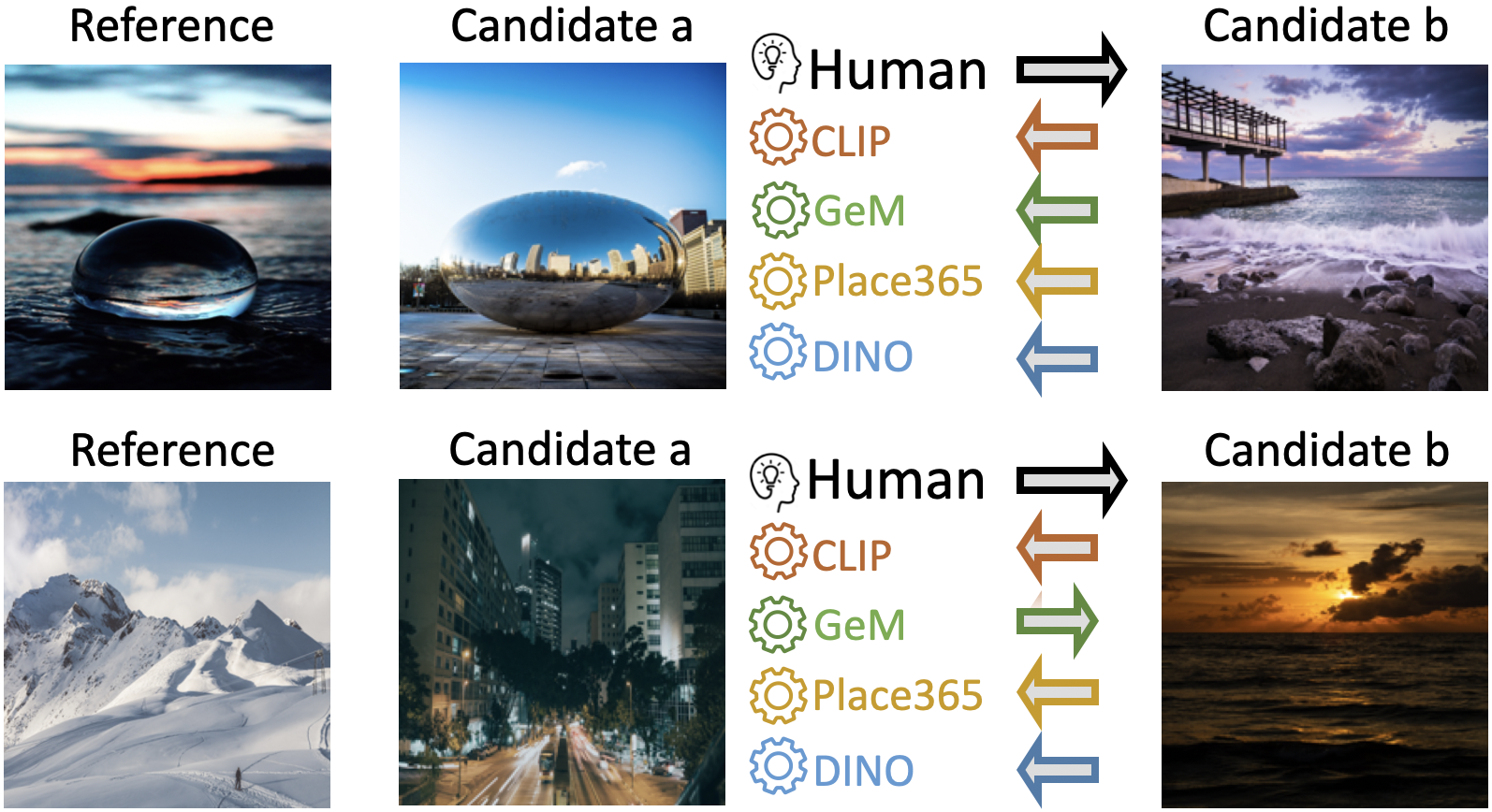}
    \caption{Each arrow points to the candidate which is considered closer to the reference by the model(s) or human annotation. Visual similarity scores computed by deep models are not always aligned with human annotations. All data, annotations, and source code used for this work can be found in \url{https://github.com/Zukang-Liao/Context-Sensitive-Image-Similarity}}
    \label{fig:disagree}
\end{figure}

\begin{figure*}
    \centering
    \includegraphics[height=64mm]{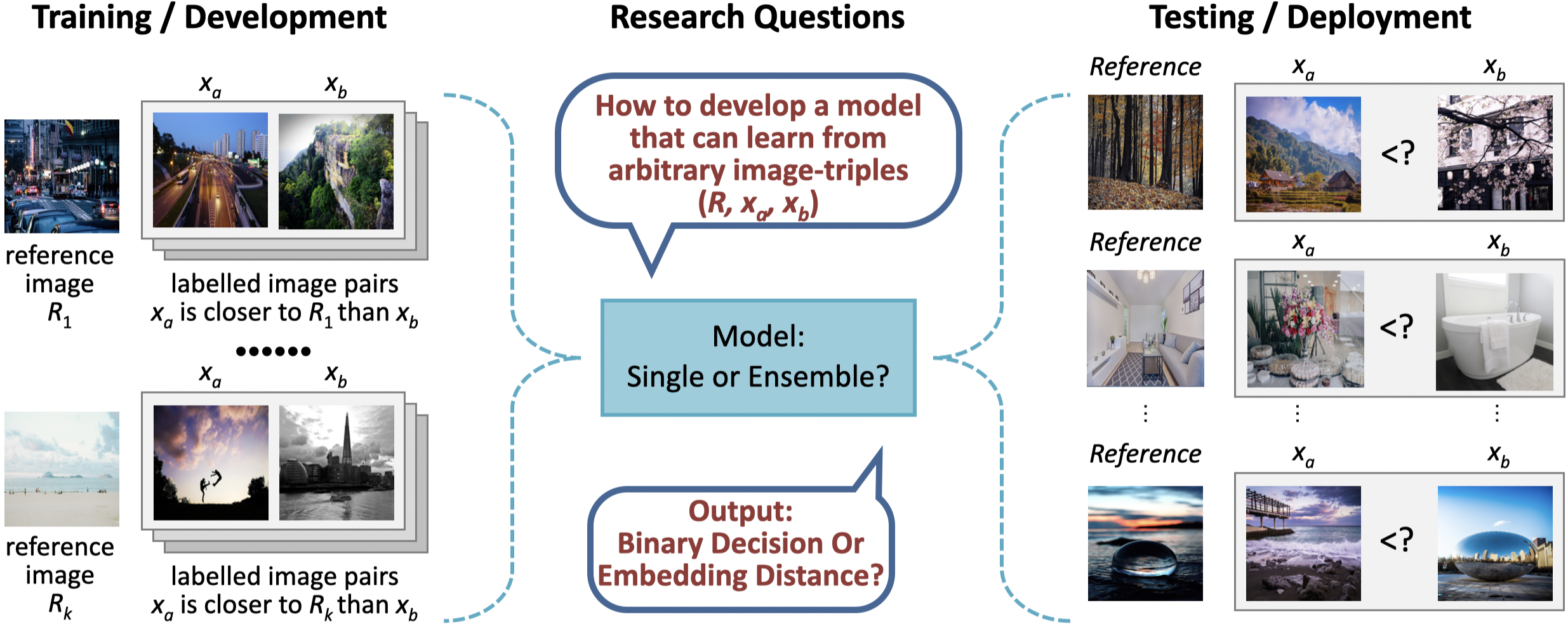}
    \caption{Given a training set of random triples that are annotated which candidate is semantically closer to the reference, can a model learn from the training data and predict correctly for unseen triples (i.e., unseen reference images and unseen candidates)?}
    \label{fig:problem_statement}
\end{figure*}

\section{Introduction}
\label{sec:intro}

Similarity between images, which has been studied for decades, is crucial for various computer vision tasks, e.g., content-based retrieval \cite{image_matching_survey} and image recognition \cite{survey_shafiq2022deep}.
In recent years, deep embeddings have become available in metrics or models for estimating image similarity, especially those machine-learned (ML) models, e.g., through metric learning \cite{metric_learning_survey} or contrastive self-supervised learning \cite{contrastive}. However, as shown in Figure \ref{fig:disagree}, deep embeddings are not always aligned with human annotations in terms of judging semantic similarity. Moreover, while similarity scores are typically of numerical values (e.g., 0.45), there is no easy way to obtain such values as ground truth data for training or testing, making it difficult to improve or compare the performance on the task of semantic similarity between images for those deep models.

Humans' perception of image similarity is often context-sensitive (CS). In some labelling processes, binary scores were assigned to image pairs in relation to a reference image (i.e., a context), i.e., is $A$ more similar to $R$ than $B$. Such labelling processes have been shown to be more consistent and objective, and have been used in image retrieval \cite{scene_graph_aaai}, face recognition \cite{schroff2015facenet}, and evaluation of synthesized images \cite{transformed_triplet_openai_cvpr}. In these areas, the existing databases are typically used to train models that can identify closely related images, i.e., either images $(R, A)$ or $(R, B)$ are very similar. For the general problem of image similarity, $A$ and $B$ can both be unrelated to $R$, as exemplified in Figure \ref{fig:disagree}. Ideally, one might wish to have a vast number of image triples $(R, A, B)$ randomly selected from an image domain $\mathbb{D}$. However, it would be costly to label these triples. Therefore, due to the gigantic data space and the limited amount of labelled data, directly fine-tuning deep models might not be effective (see Section \ref{sec:exp_global}).



In this work, we considered an alternative approach, with which we selected only a small set of $K$ reference images and the labelling effort ensured adequate sampling of $(A, B)$ in the context of each selected reference image $R_i, \;i=1..K$, as illustrated on the left of Figure \ref{fig:problem_statement}. Each $R_i$ group of labelled triples is referred to as a context-sensitive (CS) data cluster. We then obtained $K$ context-sensitive (CS) models, each of which was fine-tuned on one CS data cluster (w.r.t. a reference image $R_i$). Our experiments show that these CS models are able to improve the performance only when unseen triples contain reference images that are similar to $R_i$, e.g., when they are both flowers. We refer to such improvement as \textbf{\emph{local improvement}}, whilst the improvement on the entire dataset as \textbf{\emph{global improvement}}. We show how the performance of these CS models gradually improves locally but not globally during fine-tuning in Section \ref{sec:train_vis}. 
To fully utilize the advantage of each CS model and improve the performance on the entire dataset, we introduce two different ways to build ensemble models. To consolidate our proposed method, we compared our ensemble models with 1) existing deep embeddings, e.g., CLIP~\cite{clip} and DINO~\cite{dino}, 2) individual CS models, 3) models directly fine-tuned on the entire dataset where all labelled triples $(R, A, B)$ are amalgamated, and 4) the elementary ensemble models, e.g., majority voting. 
Our testing demonstrates that it is feasible to use CS data to develop models with little or very low context sensitivity, providing an efficient and effective approach for sampling image triples in the vast data space $\mathbb{D}^3$.

\textbf{Contributions.} In summary, (1) we revisit the problem of semantic similarity between images, and introduce a dataset with 30k labelled triples, facilitating the improvement and comparisons on the task of image semantic similarity, (2) we evaluate and compare the performance of existing deep embeddings, e.g., ViT or CLIP, and image retrieval models/algorithms, e.g., CVNet\cite{cvnet}, on the collected dataset, (3) we found fine-tuning directly on the collected dataset not effective due to the huge data space and limited amount of labelled data, (4) we found by fixing the reference image $R$, our CS models are able to improve the performance locally (when unseen reference images are similar to $R$, e.g., when they are both mountains), but not globally, (5) we provide two novel methods for constructing ensemble models using our CS models to improve the performance globally, and (6) we conduct extensive experiments to compare the proposed approach with existing methods and some more conventional solutions, and we show that the proposed method is efficient and effective when data sampling is sparse and labelling resource is limited.

\begin{table*}[t]
\centering
\caption{Comparison with existing datasets of triples}
\begin{adjustbox}{width=170mm}
\begin{tabular}{ccccccc}
\hline
Dataset & Input Type & Size  & Annotators & Data Source & Candidate Restriction & Random Candidates \\ \hline
Yoon et al.\cite{scene_graph_aaai} & Images & 1,752 & 5.7 & Visual Genome\cite{db_visual_genome}, MS-COCO\cite{db_mscoco} & Similar to the reference & \xmark \\ \hline
BAPPS (real-algo)\cite{transformed_triplet_openai_cvpr} & 64x64 Patches & 26.9k & 2.6 & MIT-Adobe5k\cite{fivek}, RAISE1k\cite{db_raise1k} & Distorted from the reference & \xmark \\ \hline
NIGHTS\cite{dreamsim} & Images    & 20k & 7.1 & Diffusion\cite{stable_diffuse}-synthesized & Synthesized from the reference & \xmark \\ \hline
CoSIS (Ours) & Images & 30k & 3 & BG20k\cite{bg20k} & No Restriction & \cmark  \\ \hline
\end{tabular}
\end{adjustbox}
\label{table:dbs}
\end{table*}

\section{Related Work}
In the literature, prevalent feature extractors, such as histograms of gradient/color or local binary patterns mainly focus on visual attributes of images with semantic information often being overlooked. For this reason, Wang et al. \cite{SDML} introduced SDML that utilized geometric mean with normalized divergences to balance inter-class divergence. Franzoni et al. \cite{random_pairs_FSKD} combined different distance measures, e.g., wordNet \cite{miller1995wordnet}, Google similarity \cite{google_similarity_distance} and tested their method on 520 random pairs collected from Flickr. Similarly, Deselaers et al. \cite{visual_semantic_cvpr} studied the relationship between visual and semantic similarity on ImageNet and they introduced a new distance metric which was shown effective on image classification tasks. Zhang et al. \cite{DeepEMD} introduced a differential earth mover’s distance (DeepEMD) and their method was proven effective on various image classification tasks under a k-shot setting. However, unlike context-based similarity, traditional scores are not consistent among different metrics and do not always have physical interoperability. Additionally, to our best knowledge, existing datasets containing triples, where two candidates can be both different from the reference are all relatively small, e.g., 520 labelled pairs \cite{random_pairs_FSKD}, or 1.7k labelled triples \cite{scene_graph_aaai}. Therefore, it is necessary to revisit the context-based similarity problem and provide a relatively larger dataset.

The BAPPS dataset\cite{transformed_triplet_openai_cvpr} consists of 26.9k triples of reference and two distorted candidates (64x64 patches). They provide the two alternative forced choice (2AFC) similarity annotations for the triples. Similarly, DreamSim\cite{dreamsim} provided 20k triples of reference and two synthesized candidates (images). D'Innocente et al. \cite{fashion_db_cvprw} provided 10,805 triples of women’s dress images. Yoon et al. \cite{scene_graph_aaai} ordered 1,752 triples of random images and defined a metric to evaluate the performance of image retrieval models. However, all the existing triples are carefully selected or synthesized. Therefore, at least one of the two candidates is noticeably similar to or almost the same as the reference image. In this work, we extend the study of image similarity to arbitrarily sampled candidates.

For image similarity, the data space is gigantic. Therefore, Wray et al. \cite{cvpr_proxy} used proxies to largely reduce the labour cost of annotation. Similarly, Movshovitz-Attias et al. \cite{iccv_proxy} used static and dynamic proxies to improve models' performance on image retrieval and clustering tasks. Given an anchor image and a smaller subset of data points (candidates), they defined the proxy as the one with minimum distance to the anchor image. This way, they showed that the loss over proxies is a tight upper bound of the original one. Aziere et al. \cite{cvpr_ensemble_proxy} trained an ensemble of CNNs using hard proxies to compute manifold similarity between images and their method was proven effective for image retrieval tasks. Similarly, Sanakoyeu et al. \cite{divid_conquer_cvpr} introduced a divide and conquer training strategy which divided the embedding space into multiple sub-spaces and then assigned each training data object a learner according to which sub-space the training data object was located.

\section{Dataset}
\label{sec:dataset}
As part of this work, we provide a new image similarity dataset (CoSIS), which currently consists of 30k labelled triples. The CoSIS dataset has 8k context-sensitive (CS) triples, which are divided into eight CS training sets (1k each) namely \emph{Indoor}, \emph{City}, \emph{Ocean}, \emph{Field}, \emph{Mountain}, \emph{Forest}, \emph{Flower}, and \emph{Abstract}. The CoSIS also contains 22k context-convolute (CC) triples, which are divided into two subsets, a validation set (12k) and a testing set (10k) for evaluating all models in an unbiased manner. 
As shown in Table \ref{table:dbs}, unlike existing datasets of triples in the literature, e.g., BAPPS \cite{transformed_triplet_openai_cvpr}, in both CS and CC portions of CoSIS, the two candidates $x_a$ and $x_b$ are selected randomly. Therefore, there is no guarantee that any of the candidates is semantically similar to or the same as the reference image $x_r$. The images of the triples are from the BG20k dataset \cite{bg20k}, which consists of 20k background images. Hence the data space of the triples is of the size of $\|\mathbb{D}^3\| = (20k)^3$.


\textbf{Two-Alternative Forced-Choice (2AFC).} For each of the 30k triples $(x_r, x_a, x_b)$, we provide binary annotation: $-1$ if $x_a$ is considered closer to $x_r$, and $1$ otherwise.


Each triple is labelled by three annotators. Among the three annotators, our inter-rater reliability score is \emph{0.947}, which is higher than most cognitive tasks, e.g., emotion detection and many NLP tasks \cite{irr}.
In some cases, we discarded triples when: (i) annotators considered two  
candidates $x_a$ and $x_b$ were very similar and equally distanced from the reference (e.g., they are both snowy mountains), and
(ii) when both $x_a$ and $x_b$ are totally irrelevant to the reference image $x_r$ (e.g., a desert and an ocean are both almost completely irrelevant to a kitchen).
With random selection, cases of (i) are relatively rare ($\leq4\%$), while cases of (ii) are more common (around $14\%$).




\textbf{Context-Sensitive (CS) Training Sets.}
For each representative reference image $R_i$, we collect 1,000 triples where the two candidates $x_a$ and $x_b$ are randomly selected. We denote it as a CS dataset $\mathbb{T}_{\text{CS}_i}$. CoSIS currently has eight CS datasets based on the eight categories defined in \cite{ref_img_categories}. The eight representative reference images are: Indoor (\emph{\#715}), City (\emph{\#2723}), Ocean (\emph{\#389}), Field (\emph{\#1673}), Mountain (\emph{\#1006}), Forest (\emph{\#254}), Flower (\emph{\#2352}), and Abstract (\emph{\#667}).
When training and evaluating each CS model on a CS dataset $D_{\text{CS}_i}$, we split the 1,000 triples into 667 for training and 333 for validation.

\textbf{Context-Convolute (CC) Data.}
These randomly selected triples were labelled for aiding the analytical ensemble strategies and evaluating fairly the performance of all models concerned. The CC data set has 22,008 triples with 2,330 unique reference images. Three images in each triple are randomly selected, and each unique reference image has at least 9 labelled triples. Therefore, the testing results for each single reference image are reasonably statistically significant. We further split the 22k random triples into a validation set (12,006 triples with 1,320 unique reference images) and a testing set (10,002 triples with 1,010 unique reference images). The validation set is used to construct ensemble models and directly fine-tune deep models (which is not effective) for comparisons, while the testing set is used for comparing the global performance of all models/algorithms concerned. Note that the testing set does not overlap with either the validation set or any CS training set, in terms of both reference images and candidate images.

\textbf{Data Cleaning.} Among the three annotators, when there were disagreements, we used majority votes as the final labels. In the original labelled triples, there were some loops (e.g., with reference $R$, $A$ is closer than $B$, $B$ is closer than $C$, $C$ is closer than $A$). We found only 0.11\% triples that were in at least a loop, and the longest loop involved four candidates. These triples were manually removed.

\begin{figure}
    \centering
    \includegraphics[height=95mm]{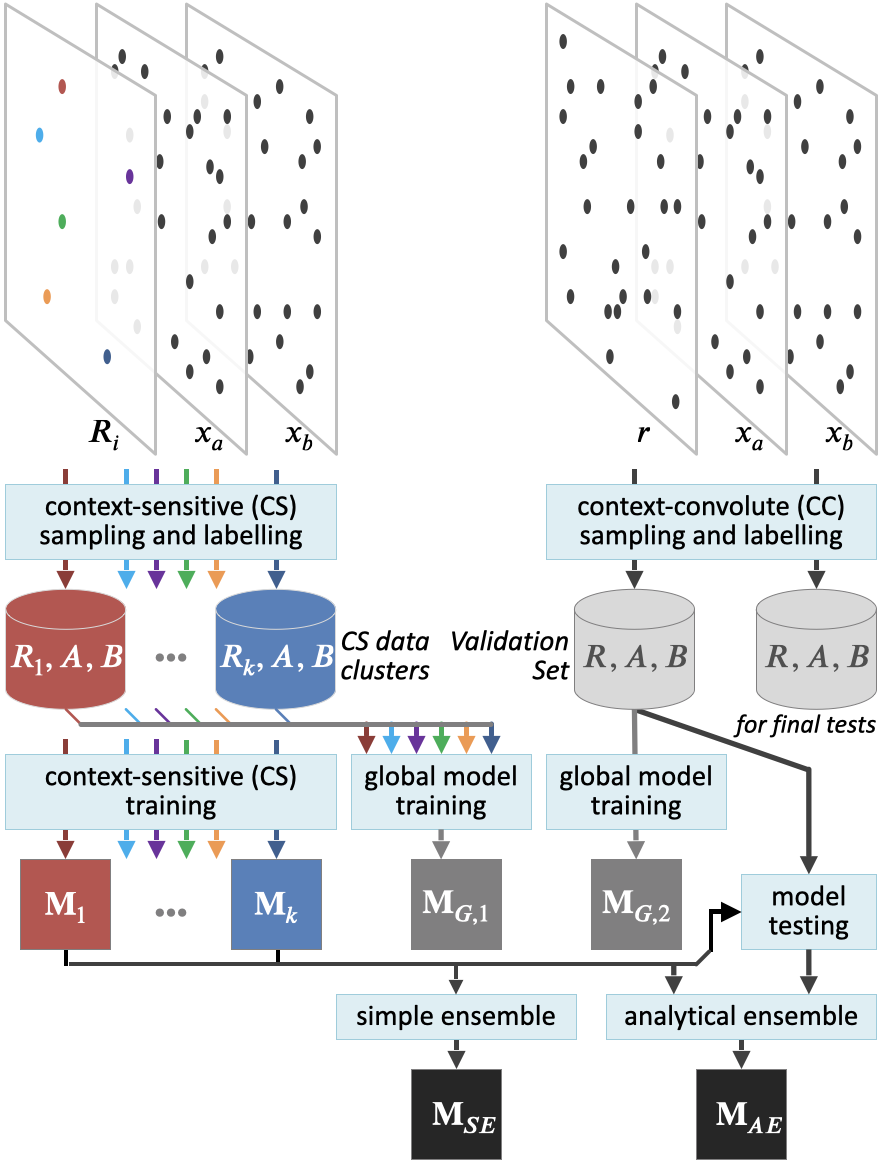}
    \caption{Workflow overview: each CS-model is trained on a CS data cluster. An analytical ensemble model is obtained based on the performance of each CS-model on the validation set. We also train global models using amalgamated data from the validation set and CS clusters for comparisons.}
    \label{fig:workflow}
\end{figure}
\section{Methodology}
\label{sec:method}
As discussed earlier, the data space of triples $(R, A, B)$ is huge. Since human intelligence is typically developed in a context-sensitive manner (e.g., most people grew up in one small region), we explore a methodology for developing similarity models based on context-sensitive (CS) learning.
As illustrated in Figure \ref{fig:workflow}, we selected several representative reference images, $R_i, i=1..K$, and for each $R_i$, we labelled $N_i$ image pairs $(A, B)$ in relation to $R_i$. We fine-tune each CS model on one of the CS clusters.
We then use the validation set of the context-convolute (CC) dataset (triples with random reference images) to conduct testing/analysis and construct ensemble models from the $K$ CS models. Finally, we test and compare all models concerned on the testing set of the CC dataset.

Due to the limited size of the annotated dataset, in this work, we focus on 1) showing that directly fine-tuning on the entire dataset is not effective (see Section \ref{sec:exp_global}), 2) demonstrating that by fixing one reference image ($R_i$) to form a CS data cluster, our CS models are able to improve local performance, i.e., when unseen reference images are similar to $R_i$, and 3) addressing the problem of lack of labelled data by constructing ensemble of our CS models to improve global performance. We leave the studies on more advanced triplet loss and more advanced deep metric training paradigms for future work when more annotated data are available.
In the following subsections, we detail the fine-tuning of CS models and the construction of ensemble models.

\subsection{Fine-tuning Models}
%
We used a simple paradigm to fine-tune our CS models on each CS data cluster. And for the purpose of comparative evaluation, we also used the same approach to fine-tune two global models, $M_{G, 1}$ and $M_{G, 2}$. The former is trained on a mixture of CS data (fixing reference images), while the latter is on the validation set of the CC data (triples with random reference images) as shown in Figure \ref{fig:workflow}.

\begin{figure}
    \centering
    \includegraphics[height=32mm]{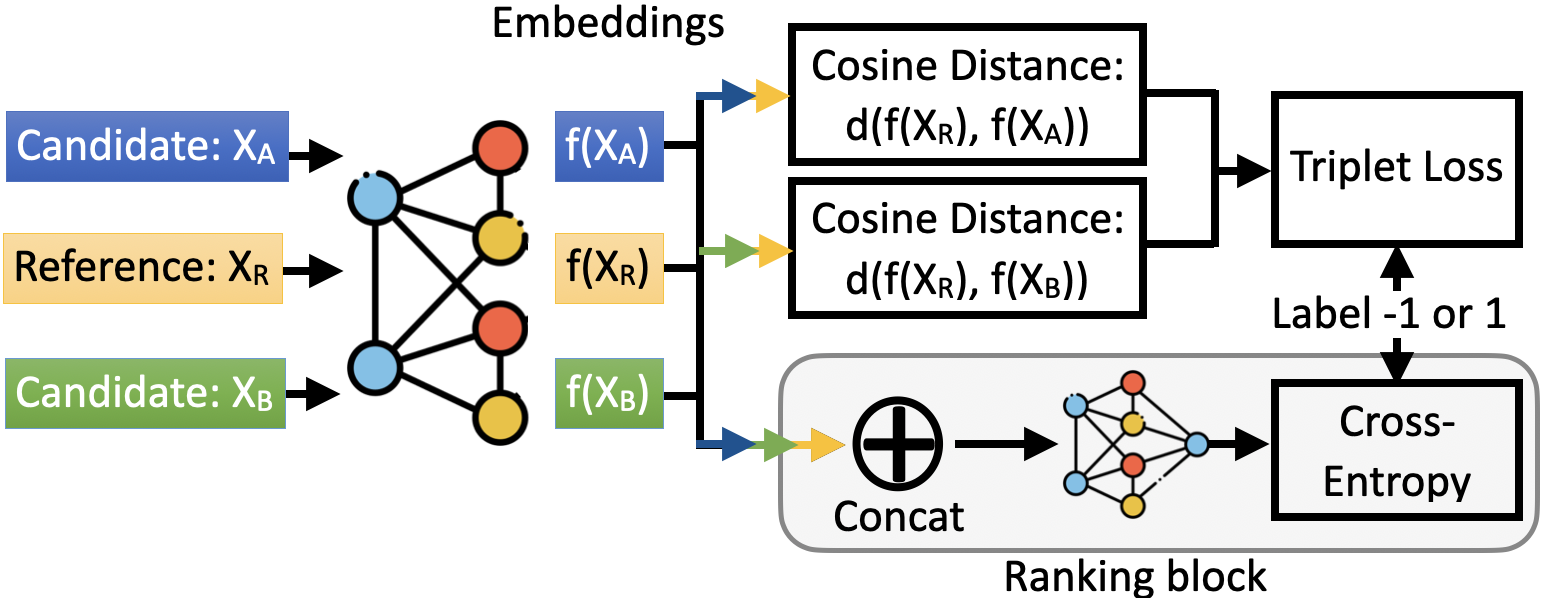}
    \caption{To train each CS model, we concatenate the embeddings and train a small ranking block to conduct binary classification. The cross-entropy loss of the ranking block, triplet loss, and LoRA \cite{lora} are used to assist in fine-tuning the backbone.}
    \label{fig:trainarch}
\end{figure}

\textbf{Training Paradigm.} As shown in Figure \ref{fig:trainarch}, we use a standard training procedure with both triplet loss and cross-entropy loss for the ranking block (binary classifier). The ranking block is helpful when the data is sparse or when the backbone is not a large model, e.g., resnet18. When the backbone is large, e.g., ViT, Lora \cite{lora} is used to reduce the number of trainable weights. 

\textbf{Context-based Triplet Loss.}
The $(R, A, B)$ triples that we used in the work are conceptually similar to the traditional contrastive / triplet loss setting (anchor, positive, negative) in deep metric learning. However, instead of pulling positive examples closer to the anchor whilst pushing negative examples away from the anchor, the selection of the two candidates $A$ and $B$ is random. One can switch $A$ and $B$ or flip the annotation for similarity augmentation. Therefore, it is not always appropriate to push $A$ or pull $B$. Formally, we define the triplet loss function as:
\begin{gather*}
    L_{\text{diff}}:=\bigl[ d\bigl(f(x_\text{ref}), f(x_a) \bigr) - d\bigr(f(x_\text{ref}), f(x_b)\bigr) \bigr] \times y\\
    L_{\text{triplet}}:=\max(margin-L_{\text{diff}}, \; 0)
\end{gather*}
where $f$ represents the backbone of an ML model ($M$), $f(x)$ denotes the embedding of an input $x$ to $M$, $d$ is a traditional distance function between two embeddings (e.g., cosine distance), and $y$ is the annotated similarity label of the triple $(x_r, x_a, x_b)$ that also controls the sign of the loss function. 

\subsection{Ensemble Strategies}

Our Ensemble Strategies are constructed based on the performance of each CS model. In this section, we first detail how we analyse the testing results of each model on the validation set of our CC dataset (triples with random reference images), as well as how we construct ensemble models based on the analysis.

\begin{figure}
    \centering
    \includegraphics[height=52.5mm]{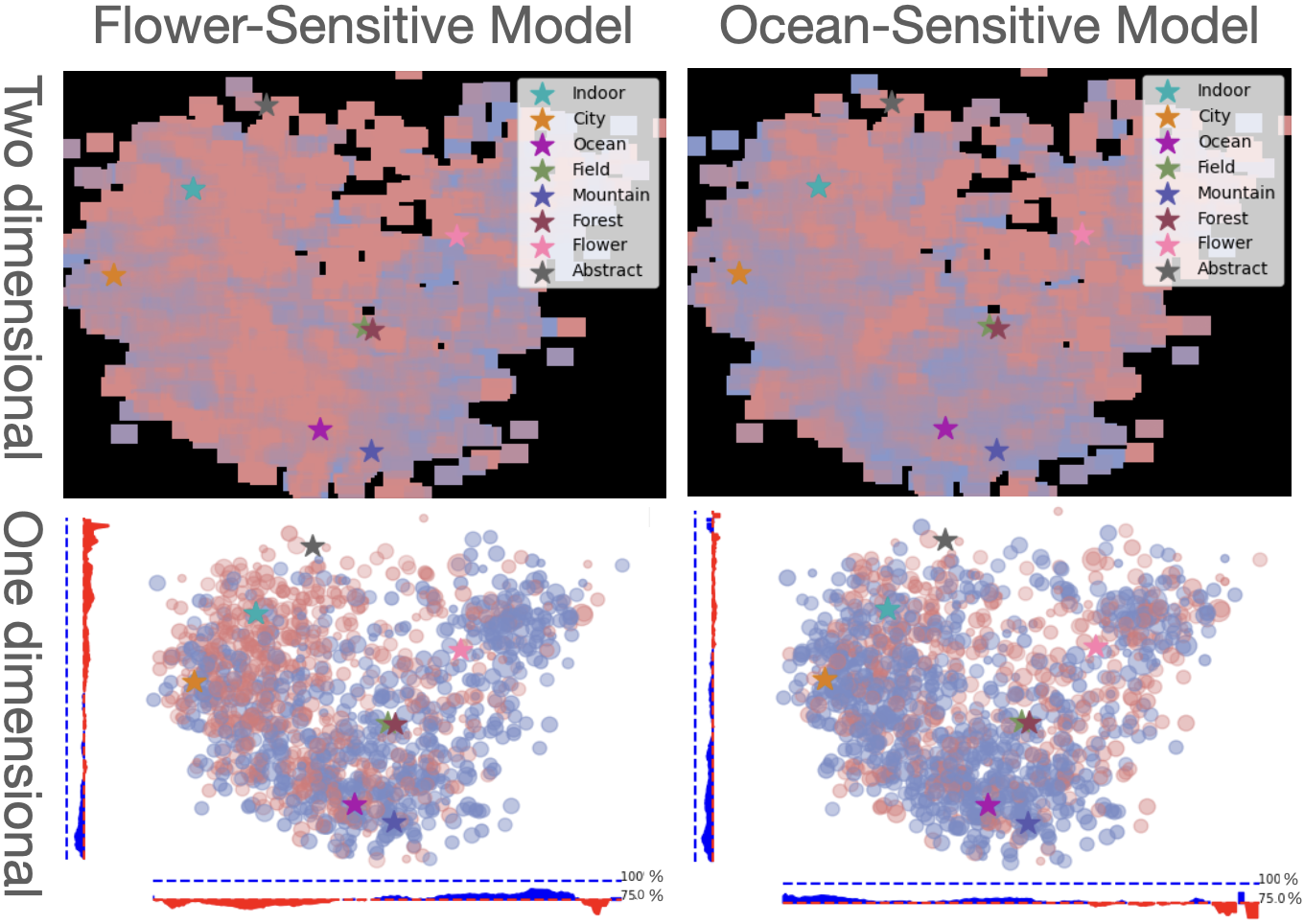}
    \caption{Ensemble Approach (PCA): for all triples $(R_i, x_a, x_b)$ sharing the same reference image $R_i$, we compute an accuracy score from each model. We visualize the accuracy scores of the $| \mathbb{T}_V |$ reference images in our validation set using PCA or tSNE. Different models perform well in different areas. An ensemble method can be obtained based on the scatter plots.}
    \label{fig:ensemble}
\end{figure}

\textbf{Context-Convolute (CC) Testing.} While one may train a set of CS models, one would like to use these models to construct an ensemble that can be applied to other contexts, i.e., when the testing triples include unseen reference images. In our CC dataset, we include multiple triples with $\|\mathbb{T}_V \|$ random reference images, and for each random reference image, there are at least nine labelled triples (see Section \ref{sec:dataset}). Therefore, in addition to a global accuracy score, we can report the testing results on the CC dataset based on individual reference images.
For all triples $(R_i, x_a, x_b)$ sharing the same reference image $R_i$, and each CS model $\mathrm{M}_i, i=1..K$, the testing yields a correctness indicator. The total number of such indicators is $\|\mathbb{T}_V \| \times K$. The testing can also result in additional information, such as confusion matrix and uncertainty or confidence values. The CC testing results can inform the construction of ensemble models. 







\textbf{Feature-based Analysis and Specialization.}
Given a large set of images, one can extract $l$ features, which define an $l$-D feature space of the images. These features can be the results of dimensionality reduction techniques such as PCA and t-SNE as well as hand-crafted feature extraction algorithms. Each image can thus be encoded as an $l$-D feature vector $\Theta = [\theta_1, \theta_2, ..., \theta_l]$.

For an arbitrary reference image $R_i$, its feature vector $\Theta_r$ determines its $l$-D coordinates in the feature space. When all triples $(R_i, x_a, x_b)$ sharing the same reference image $R_i$ are tested against a CS model $\mathrm{M}_i$, the correctness indicator can be considered as a sample of a correctness manifold at position $\Theta$. With $K$ CS models, we have $K$ such manifolds based on correctness indicators. The testing of a CS model $\mathrm{M}_i$ on the validation set provides us with a way to establish an approximate model of the manifold that can be used to predict the correctness of applying $\mathrm{M}_i$ to a previously-unseen image triple as shown in Figure \ref{fig:ensemble}.

\textbf{Ensemble based on Credibility Maps.} An $l$-D \emph{credibility map} of a model $\mathrm{M}_i$ is a discrete partition of the $l$-D feature space into a number of $l$-D cells, and each cell stores a value (or values) indicating the probability of $\mathrm{M}_i$ to be correct when it applies to any image triples $(R_i, x_a, x_b)$ where the feature vector of $R_i$ falls into the cell. Figure \ref{fig:ensemble} illustrates such credibility maps in 1D and 2D sub-spaces. The two plots above show the 2D credibility maps of two CS models fine-tuned on the flower data cluster and ocean data cluster respectively. Each of the 2D manifolds is sampled on 12k triples with 1,320 different reference images, and the 2D feature subspace is partitioned into $200^2$ cells. The two images below include four line plots representing four 1D credibility maps. Each line plot results from the projection of a set of testing results.
From Figure \ref{fig:ensemble}, we can observe that these credibility maps can provide useful information about the past and potential performance of different CS models in different parts of the feature space.

One ensemble strategy is to determine, for any image triple $(R_i, x_a, x_b)$, how much each CS model should contribute to the decision. The feature vector of the reference image $R_i$ is used to look up relevant cells in one or more credibility maps. Likely partitioning a high-dimensional feature space will result in many empty cells. A practical solution is for an ensemble algorithm to consult several low-dimensional credibility maps for each CS model and aggregate the credibility values into a single credibility score per CS model. The scores for different CS models can then be used to determine the contribution of each CS Model in the final decision specifically for the image triple(s) $(R_i, x_a, x_b)$.


\begin{figure}
    \centering
    \includegraphics[height=52.5mm]{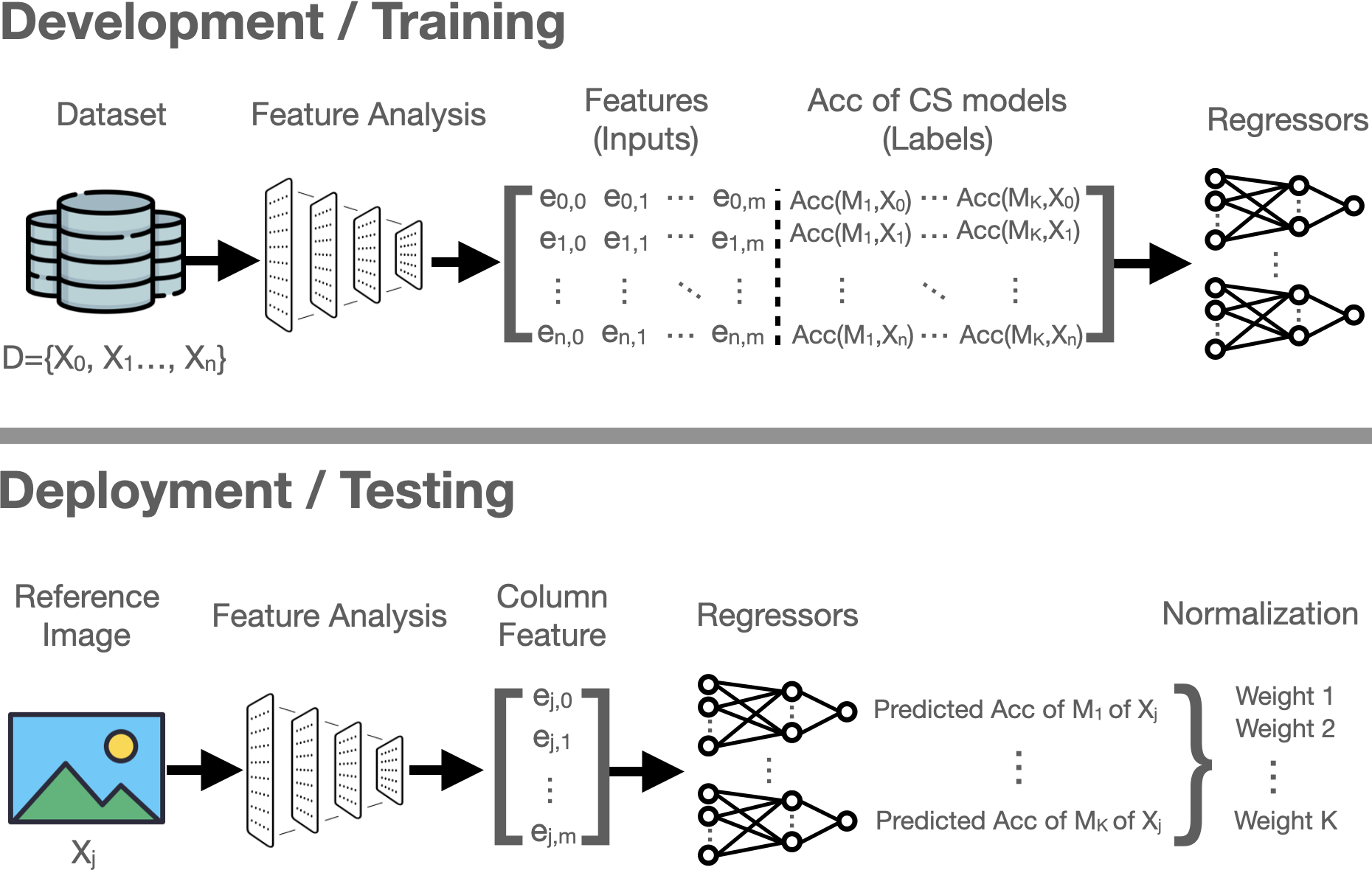}
    \caption{Ensemble Approach (MLP): The input of the MLP regressors is the features of a reference image $R_i$, and the outputs are the predicted accuracy score of each CS model on all triples $(R_i, x_a, x_b)$ sharing the same reference image $R_i$. For any previously unseen triple, the outputs can be used as the tailored ensemble weights of the CS models.}
    \label{fig:ensemble_mlp}
\end{figure}

\begin{table*}[h]
\centering
\caption{Local Performance of existing supervised and self-supervised models on different Context-Sensitive testing clusters.}
\begin{adjustbox}{width=170mm}
\begin{tabular}{|lccccccccc|}
\hline
\multicolumn{10}{|c|}{\textit{\textbf{Traditional Similarity Metrics and Deep Image Retrieval Models}}}  \\ \hline
\multicolumn{1}{|m{36mm}|}{Reference Image} & \multicolumn{1}{c|}{\#Indoor}  & \multicolumn{1}{c|}{\#City} & \multicolumn{1}{c|}{\#Ocean}  & \multicolumn{1}{c|}{\#Field} & \multicolumn{1}{c|}{\#Mountain} & \multicolumn{1}{c|}{\#Forest}  & \multicolumn{1}{c|}{\#Flower} & \multicolumn{1}{c|}{\#Abstract}  & \multicolumn{1}{c|}{Average} \\ \hline
\multicolumn{1}{|l|}{CVNet\cite{cvnet}}   & \multicolumn{1}{c|}{67.4\%} & \multicolumn{1}{c|}{58.8\%} & \multicolumn{1}{c|}{71.6\%} & \multicolumn{1}{c|}{52.8\%} & \multicolumn{1}{c|}{62.3\%} & \multicolumn{1}{c|}{56.9\%} & \multicolumn{1}{c|}{58.9\%} & \multicolumn{1}{c|}{64.6\%} & 61.6\%       \\ \hline
\multicolumn{1}{|l|}{HesAff SIFT+SP\cite{Hessian_sift}}   & \multicolumn{1}{c|}{66.8\%} & \multicolumn{1}{c|}{60.0\%} & \multicolumn{1}{c|}{55.1\%} & \multicolumn{1}{c|}{41.8\%} & \multicolumn{1}{c|}{53.3\%} & \multicolumn{1}{c|}{37.7\%} & \multicolumn{1}{c|}{54.4\%} & \multicolumn{1}{c|}{59.2\%} & 53.5\%       \\ \hline
\multicolumn{1}{|l|}{GeM-ResNet50\cite{RTC18}}     & \multicolumn{1}{c|}{73.7\%} & \multicolumn{1}{c|}{77.6\%} & \multicolumn{1}{c|}{75.1\%} & \multicolumn{1}{c|}{71.0\%} & \multicolumn{1}{c|}{68.0\%} & \multicolumn{1}{c|}{77.2\%} & \multicolumn{1}{c|}{73.9\%} & \multicolumn{1}{c|}{55.9\%} & 71.6\%  \\ \hline
\multicolumn{1}{|l|}{SfM-ResNet50\cite{RTC16}}    & \multicolumn{1}{c|}{81.9\%} & \multicolumn{1}{c|}{81.6\%} & \multicolumn{1}{c|}{85.8\%} & \multicolumn{1}{c|}{69.5\%} & \multicolumn{1}{c|}{83.3\%} & \multicolumn{1}{c|}{74.8\%} & \multicolumn{1}{c|}{73.6\%} & \multicolumn{1}{c|}{69.8\%} & 77.5\%       \\ \hline

\noalign{\medskip}\hline

\multicolumn{10}{|c|}{\textit{\textbf{Large Self-Supervised Models Trained on Large Datasets}}} \\ \hline
\multicolumn{1}{|m{36mm}|}{Reference Image} & \multicolumn{1}{c|}{\#Indoor}  & \multicolumn{1}{c|}{\#City} & \multicolumn{1}{c|}{\#Ocean}  & \multicolumn{1}{c|}{\#Field} & \multicolumn{1}{c|}{\#Mountain} & \multicolumn{1}{c|}{\#Forest}  & \multicolumn{1}{c|}{\#Flower} & \multicolumn{1}{c|}{\#Abstract}  & \multicolumn{1}{c|}{Average} \\ \hline
\multicolumn{1}{|l|}{DINO\cite{dino}-ResNet50} & \multicolumn{1}{c|}{69.2\%} & \multicolumn{1}{c|}{81.2\%} & \multicolumn{1}{c|}{76.6\%} & \multicolumn{1}{c|}{70.7\%} & \multicolumn{1}{c|}{81.7\%} & \multicolumn{1}{c|}{74.6\%} & \multicolumn{1}{c|}{77.8\%} & \multicolumn{1}{c|}{49.2\%} & \multicolumn{1}{c|}{72.6\%} \\ \hline
\multicolumn{1}{|l|}{DINO\cite{dino}-ViT-B16} & \multicolumn{1}{c|}{74.6\%} & \multicolumn{1}{c|}{83.3\%} & \multicolumn{1}{c|}{83.8\%} & \multicolumn{1}{c|}{69.0\%} & \multicolumn{1}{c|}{87.7\%} & \multicolumn{1}{c|}{72.5\%} & \multicolumn{1}{c|}{75.4\%} & \multicolumn{1}{c|}{70.3\%} & \multicolumn{1}{c|}{77.1\%} \\ \hline
\multicolumn{1}{|l|}{CLIP\cite{clip}-ViT-B32} & \multicolumn{1}{c|}{70.1\%} & \multicolumn{1}{c|}{78.8\%} & \multicolumn{1}{c|}{82.9\%} & \multicolumn{1}{c|}{68.7\%} & \multicolumn{1}{c|}{85.6\%} & \multicolumn{1}{c|}{79.0\%} & \multicolumn{1}{c|}{77.2\%} & \multicolumn{1}{c|}{80.8\%} & \multicolumn{1}{c|}{77.9\%} \\ \hline

\noalign{\medskip}\hline
\multicolumn{10}{|c|}{\textit{\textbf{Embedding Distances Based on Backbones of Supervised Models Trained on Large Datasets}}} \\ \hline
\multicolumn{1}{|m{36mm}|}{Reference Image} & \multicolumn{1}{c|}{\#Indoor}  & \multicolumn{1}{c|}{\#City} & \multicolumn{1}{c|}{\#Ocean}  & \multicolumn{1}{c|}{\#Field} & \multicolumn{1}{c|}{\#Mountain} & \multicolumn{1}{c|}{\#Forest}  & \multicolumn{1}{c|}{\#Flower} & \multicolumn{1}{c|}{\#Abstract}  & \multicolumn{1}{c|}{Average} \\ \hline
\multicolumn{1}{|l|}{ResNet18-Place365\cite{place365}} & \multicolumn{1}{c|}{61.1\%} & \multicolumn{1}{c|}{78.2\%} & \multicolumn{1}{c|}{78.1\%} & \multicolumn{1}{c|}{69.6\%} & \multicolumn{1}{c|}{83.8\%} & \multicolumn{1}{c|}{77.5\%} & \multicolumn{1}{c|}{72.1\%} & \multicolumn{1}{c|}{72.1\%} & 74.1\% \\\hline
\multicolumn{1}{|l|}{ResNet18-ImageNet\cite{resnet}} & \multicolumn{1}{c|}{76.9\%} & \multicolumn{1}{c|}{82.4\%} & \multicolumn{1}{c|}{81.1\%} & \multicolumn{1}{c|}{66.0\%} & \multicolumn{1}{c|}{87.4\%} & \multicolumn{1}{c|}{72.2\%} & \multicolumn{1}{c|}{66.1\%} & \multicolumn{1}{c|}{67.6\%} & 75.0\% \\\hline
\multicolumn{1}{|l|}{VGG16-ImageNet\cite{vgg}} & \multicolumn{1}{c|}{81.4\%} & \multicolumn{1}{c|}{82.7\%} & \multicolumn{1}{c|}{88.9\%} & \multicolumn{1}{c|}{62.4\%} & \multicolumn{1}{c|}{88.6\%} & \multicolumn{1}{c|}{75.4\%} & \multicolumn{1}{c|}{70.6\%} & \multicolumn{1}{c|}{67.0\%} & 77.1\% \\\hline
\multicolumn{1}{|l|}{ViT-ImageNet\cite{vit}} & \multicolumn{1}{c|}{82.3\%} & \multicolumn{1}{c|}{83.1\%} & \multicolumn{1}{c|}{83.0\%} & \multicolumn{1}{c|}{76.0\%} & \multicolumn{1}{c|}{87.4\%} & \multicolumn{1}{c|}{83.8\%} & \multicolumn{1}{c|}{86.8\%} & \multicolumn{1}{c|}{72.9\%} & 81.9\% \\\hline
\end{tabular}
\end{adjustbox}
\label{table: acc_context_dataset_reviews}

\medskip
\centering
\caption{Local performance of different CS models (trained on the CS training dataset) on the corresponding testing dataset.}
\begin{adjustbox}{width=170mm}
\begin{tabular}{|lccccccccc|}
\hline

\multicolumn{10}{|c|}{\textit{\textbf{Our Context-Sensitive Models - Different Pre-trained Architectures}}}\\ \hline
\multicolumn{1}{|m{36mm}|}{Reference Image} & \multicolumn{1}{c|}{\#Indoor}  & \multicolumn{1}{c|}{\#City} & \multicolumn{1}{c|}{\#Ocean}  & \multicolumn{1}{c|}{\#Field} & \multicolumn{1}{c|}{\#Mountain} & \multicolumn{1}{c|}{\#Forest}  & \multicolumn{1}{c|}{\#Flower} & \multicolumn{1}{c|}{\#Abstract}  & \multicolumn{1}{c|}{Average} \\ \hline
\multicolumn{1}{|l|}{VGG16-ImageNet} & \multicolumn{1}{c|}{86.2\%} & \multicolumn{1}{c|}{86.6\%} & \multicolumn{1}{c|}{86.1\%} & \multicolumn{1}{c|}{\textbf{90.1\%}} & \multicolumn{1}{c|}{87.8\%} & \multicolumn{1}{c|}{76.9\%} & \multicolumn{1}{c|}{72.3\%} & \multicolumn{1}{c|}{74.5\%} & 82.6\% \\ \hline
\multicolumn{1}{|l|}{ResNet18-ImageNet} & \multicolumn{1}{c|}{\textbf{88.0\%}} & \multicolumn{1}{c|}{86.0\%} & \multicolumn{1}{c|}{88.0\%} & \multicolumn{1}{c|}{78.8\%} & \multicolumn{1}{c|}{90.4\%} & \multicolumn{1}{c|}{82.9\%} & \multicolumn{1}{c|}{85.5\%} & \multicolumn{1}{c|}{79.3\%} & \multicolumn{1}{c|}{84.8\%} \\ \hline
\multicolumn{1}{|l|}{ResNet18-Place365} & \multicolumn{1}{c|}{86.5\%} & \multicolumn{1}{c|}{\textbf{87.5\%}} & \multicolumn{1}{c|}{88.6\%} & \multicolumn{1}{c|}{88.1\%} & \multicolumn{1}{c|}{90.4\%} & \multicolumn{1}{c|}{77.8\%} & \multicolumn{1}{c|}{70.6\%} & \multicolumn{1}{c|}{\textbf{83.8\%}} & \multicolumn{1}{c|}{84.2\%} \\ \hline
\multicolumn{1}{|l|}{ViT-Lora} & \multicolumn{1}{c|}{80.2\%} & \multicolumn{1}{c|}{82.7\%} & \multicolumn{1}{c|}{\textbf{89.5\%}} & \multicolumn{1}{c|}{83.6\%} & \multicolumn{1}{c|}{\textbf{90.7\%}} & \multicolumn{1}{c|}{\textbf{86.2\%}} & \multicolumn{1}{c|}{\textbf{86.8\%}} & \multicolumn{1}{c|}{78.4\%} & \textbf{84.8\%} \\ \hline
\end{tabular}
\end{adjustbox}
\label{table: acc_context_dataset}
\end{table*}

\textbf{ML-Based Ensemble Strategy.}
The weights of the CS models can also be produced by another ML model, which is trained using the feature vector of $R_i$ as the input and the accuracy scores on all triples $(R_i, x_a, x_b)$ (sharing the same reference images $R_i$) as the correctness label.
Theoretically, the weights can be jointly learned by a large ML model with a larger dataset.
In this work, we show that, with a relatively small validation set (12k triples with 1,320 unique reference images), we can still train another simple ML model to predict the performance of each CS model. As shown in Figure \ref{fig:ensemble_mlp}, firstly, we extract features of the reference images in the validation set using a neural net (e.g., ViT), and we then use a dimensionality reduction method (e.g., PCA) to counteract the sparseness of the annotated data. In our main implementation, we used 64 dimensions. The feature of a reference image $R_i$ is fed to several multi-layer perceptrons (MLPs), each of which is trained to predict the accuracy score of one CS model on all triples with $R_i$. In the deployment process, given a random triple, the MLPs estimate the likelihood score of each CS model making a correct decision. The normalized scores are used as the weights for determining the contribution of each CS model for the specific triple.


\section{Experiment}
\label{sec:exp}
To consolidate our methodology, we conducted extensive tests. In this section, we use experimental results to show: 

\noindent\textbf{• Local improvement of CS Models.} We fine-tuned eight CS models $\mathrm{M}_i, i=1..8$, each of which was fine-tuned on the training-split of one CS data cluster (667 triples with fixed reference image $R_i$, see Section \ref{sec:dataset}). We show that our CS models are able to improve performance only when unseen triples contain reference images that are similar to $R_i$ (local improvement), but not on the entire CC testing set of 10k random triples (no global improvement).

\noindent\textbf{• Limited global performance if directly fine-tuning.} We directly fine-tuned two global models, one on the CC validation set (12k triples with random reference images), and the other on the amalgamated CS training sets (8k in total). Due to the limited amount of labelled data, these two global models do not perform well on the CC testing set (10k).

\noindent\textbf{• Effective Ensemble Models of CS models.} We analyze the performance of our CS models on the CC validation set (12k), and construct ensemble models to improve the global performance on the CC testing set (10k),

\subsection{Fine-tuning Context-Sensitive Models} 
When fine-tuning each CS model, we use the following setting: 1) learning rate: 10$^{-4}$ for ViT with LoRA, and 10$^{-5}$ for others architectures, 2) number for epochs: 25, 3) loss function: cross entropy + 0.1 $\times$ triplet loss, 4) batch size: 8, 5) optimizer: adam, 6) single image augmentation: random resized crop and horizontal flip, 7) triples augmentation: randomly swap candidate A and B, 8) all images resized to: 224$\times$224. We did not carefully tune the hyper-parameters. Due to the small size of each CS cluster, we are able to fine-tune all of our CS models on one single laptop with an Apple M1 Chip, including ViT with LoRA. The training and fine-tuning time of each CS model varies from one day to three days (with ViT and LoRA). With the limited resources and limited amount of labelled data, we are able to improve the performance on the problem of context-sensitive image similarity using the proposed methodology.

\begin{figure*}[t]
    \centering
    \includegraphics[height=49mm]{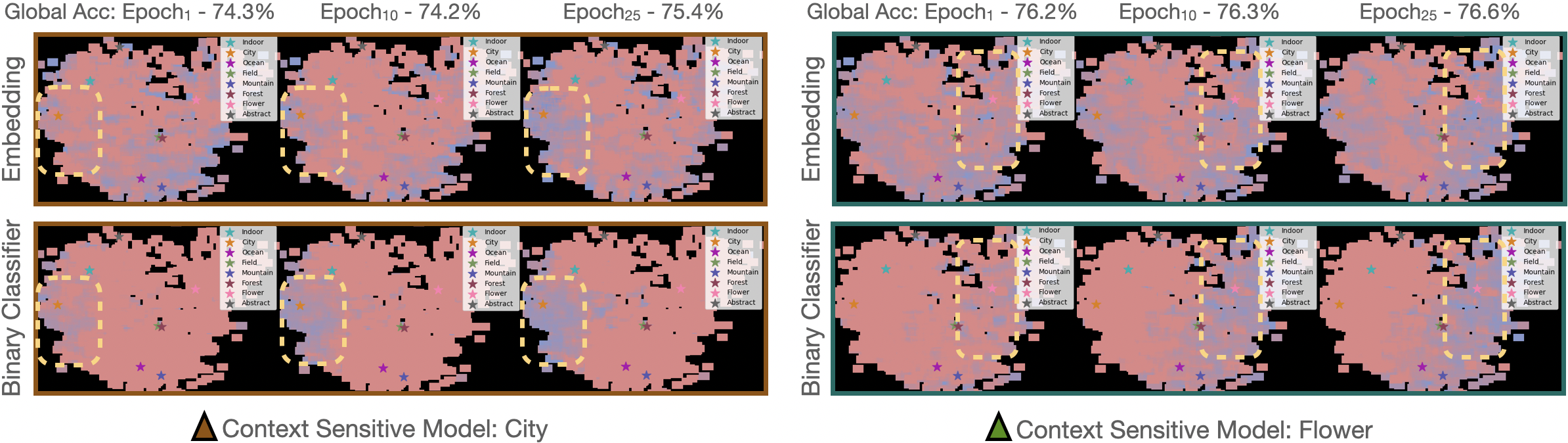}
    \caption{Visualized CS training: the local performance (highlighted areas) is improved gradually during training, whilst the global accuracy remains stable. The second row shows the changes in binary classifiers' performance from scratch. It is more noticeable that the performance in the highlighted areas is constantly improving during training. The first row shows that we can also see the same improvement in the highlighted areas when using embeddings, especially when comparing the results at the beginning and end of CS training. This shows that CS training improves local performance for both binary classifiers and embeddings. The bluer, the more accurate the CS model that is being trained, whilst red indicates accuracy $\leq$75\%.}
    \label{fig:step}
\end{figure*}

\begin{table*}[t]
\centering
\caption{No significant global improvement for CS models. As also shown in Figure \ref{fig:step}, CS models are able to achieve local improvement but not global improvement.}
\begin{adjustbox}{width=140mm}
\begin{tabular}{|l|c|c|c|c|c|c|c|c|}
\hline
Context-Sensitive Model: & Indoor & City & Ocean & Field & Mountain & Forest & Flower & Abstract \\ \hline
Accuracy on Validation Set (12k) & 77.3\% & 75.4\% & 78.3\% & 73.3\% & 79.5\% & 75.7\% & 76.6\% & 79.3\% \\ \hline
Accuracy on Testing Set (10k) & 76.9\% & 74.3\% & 78.4\% & 72.7\% & 79.1\% & 73.9\% & 73.9\% & 77.2\% \\ \hline
\end{tabular}
\end{adjustbox}
\label{table:cs_model_valtestset}
\end{table*}

\subsection{Performance of Context-Sensitive Models}
%

\textbf{Performance on the CS Training Sets.} As shown in Table \ref{table: acc_context_dataset_reviews}, using embedding distances from image retrieval models achieved around 60\%$\sim$78\%, and large self-supervised models, e.g., CLIP/DINO, and supervised models, e.g., ViT, were able to achieve 78\%$\sim$82\%. To investigate whether we can improve the performance, we fine-tuned several models of different architectures. As shown in Table \ref{table: acc_context_dataset}, as expected, the fine-tuned CS models outperform all existing methods on their corresponding CS datasets. In Table \ref{table: acc_context_dataset_more} in the appendix, we show more results on using different architectures. Due to the limited number of labelled data, compared with ResNet18 and VGGNets, larger architectures might not be effective, e.g., ResNet34 and ResNet50 reached around 60\%$\sim$65\% whilst ResNet18 and VGGNets reached around 81\%$\sim$84\% on average. Therefore, we fine-tuned larger and deeper models with LoRA\cite{lora} to boost the performance. ViT with Lora (denoted as ViT-Lora) achieved the best averaged accuracy (around 85\%), which is aligned with the most recent studies on evaluating synthesized images using labelled triples \cite{dreamsim, transformed_triplet_openai_cvpr}.

In Table \ref{table: acc_context_dataset}, we highlight the best CS model fine-tuned on each CS data cluster. These eight CS models achieved 84\%$\sim$91\%, which is significantly better than all of the existing methods in Table \ref{table: acc_context_dataset_reviews}. The results suggest that our CS models can outperform existing models locally: when triples contain seen reference image $R_i$ and unseen candidates $x_a$ and $x_b$. In Section \ref{sec:train_vis} and Figure \ref{fig:step}, we visualize the CS training procedure, and the results also show that our CS models are able to improve performance locally: when the triples contain reference images that are similar to $R_i$. 





\noindent\textbf{Performance on the CC Validation (12k) / Testing Set (10k):}
To evaluate how the selected CS models perform on triples with random unseen reference images, we run each CS model $\mathrm{M}_i$ on the CC validation and testing set. As shown in Table \ref{table:cs_model_valtestset}, the CS models achieved 73\%$\sim$79.5\%, which is lower than the results they achieved on the CS clusters. And the results are also slightly worse or similar to the existing models, e.g., ViT, as shown in Table \ref{table: ind_results}. This shows that our CS models can only improve performance locally (on triples with similar reference images) but not globally (on triples with random reference images).
%

\begin{table}[b]
\centering
\caption{Performance of global models on the testing set (10k)}
\begin{adjustbox}{width=85mm}
\begin{tabular}{|l|c|c|c|}
\hline
\diagbox[width=30mm]{Architecture}{Training Set} & \begin{tabular}[c]{@{}c@{}}No Training\\ Embedding Distance \end{tabular} & \begin{tabular}[c]{@{}c@{}}Context Training \\ Set (8k)\end{tabular} & \begin{tabular}[c]{@{}c@{}}Validation \\ Set (12k)\end{tabular} \\ \hline
VGG16 & 78.3\% & 71.4\%$\pm$4.2\%  & 77.3\%$\pm$2.0\% \\ \hline
ResNet18 & 77.7\% & 72.1\%$\pm$1.9\%  & 80.2\%$\pm$1.6\% \\ \hline
ViT-Lora & 79.9\% & 68.9\%$\pm$3.6\%  & 79.6\%$\pm$1.3\% \\ \hline
\end{tabular}
\end{adjustbox}
\label{table:global_model}
\end{table}

\subsection{Performance of Global Models}
\label{sec:exp_global}
%

One straightforward potential solution to improve global performance is to fine-tune deep models on triples with random reference images. Therefore, we fine-tuned two global models, $\mathrm{M}_{G1}$ and $\mathrm{M}_{G2}$. The former was fine-tuned with an amalgamation of the eight CS data clusters (8k triples in total), and the latter with the CC validation set (12k triples with 1,320 random reference images). Three architectures were used. Table \ref{table:global_model} shows the results of testing these models on the CC testing set (10k triples with 1,010 unique and random reference images).

\begin{figure*}[t]
    \centering
    \includegraphics[height=37.5mm]{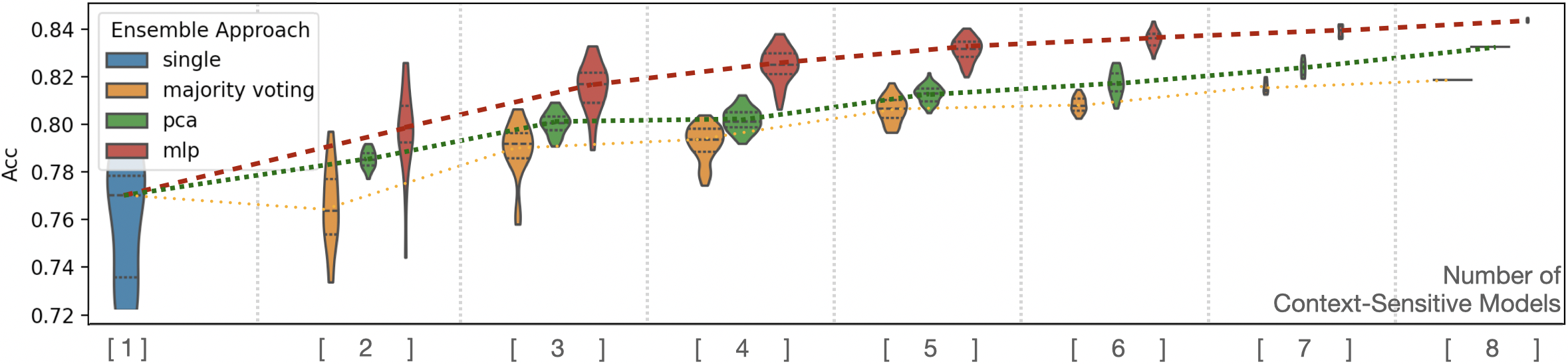}
    \caption{Y-axis: accuracy of the ensemble methods on the testing set (10k). X-axis: the number of CS models used to form the ensemble model. Experiments are run on all the combinations, e.g., when choosing two CS models, we run experiments on all of the $C_{2}^{8}$=$\frac{8!}{2!*6!}$=28 combinations. For MLP, we repeat the same experiment three times for one combination, e.g., for choosing two CS models, we run 3*$C_{2}^{8}$=84 experiments. MLP-based approach consistently performs the best. Dashed lines inside the blobs are the quartiles of the data.}
    \label{fig:number_proxies}
\end{figure*}

\begin{table*}[t]
\caption{Global performance comparisons on validation set: 12k random triples and testing set: 10k random triples}
\begin{tabular}{l|ccc|ccc}
\hline
& \multicolumn{3}{c|}{\textbf{Best Single Model}}                                                                             & \multicolumn{3}{c}{\textbf{Ensemble of CS models}}                                                    \\ \hline
Description    & \multicolumn{1}{l}{Not fine-tuned: ViT\cite{vit}} & \multicolumn{1}{l}{Globally fine-tuned} & \multicolumn{1}{l|}{Locally fine-tuned} & \multicolumn{1}{l}{Majority Voting} & \multicolumn{1}{l}{PCA (Ours)} & \multicolumn{1}{l}{MLP (Ours)} \\\hline
Validation Set & 80.4\%                                  & 80.5\%$\pm$1.2\%          & 79.5\%                                  & 82.3\%                              & \textbf{87.0\%}                & 86.6\%$\pm$0.5\%                       \\\hline
Testing Set    & 79.9\%                                  & 80.2\%$\pm$1.6\%          & 79.1\%                                  & 81.9\%                              & 83.3\%        & \textbf{84.7\%}$\pm$0.3\% \\\hline 
\end{tabular}
\label{table: ind_results}
\end{table*}

The global models trained on the CC validation set ($\mathrm{M}_{G2}$) achieved 77\%$\sim$80\% on average, which is similar to existing models, e.g., ViT: $\sim$80\% and ResNet18: $\sim$77\%, which are not trained on any of our labelled data. Additionally, fine-tuning on all of our context training sets (8k) ($\mathrm{M}_{G1}$) does not improve the performance on the testing set either. Moreover, fine-tuning one single model on all of the amalgamated context datasets led to a decrease in accuracy on the testing set (69\%$\sim$72\%) compared with their untrained counterparts (77\%$\sim$80\%). This might be caused by the sparsity of the context training set which only contains eight reference images. 

The worse performance (compared with CLIP or DINO) of these two global models shows that: directly fine-tuning on random triples does not improve the performance due to the huge data space and limited amount of labelled data. By contrast, fixing the reference image $R_i$ hugely reduces the size of data needed to fine-tune the neural nets. Therefore, each single CS model is able to outperform existing algorithms on triples with seen reference images that are similar to $R_i$ (Figure \ref{fig:step}). To boost the global performance (on random triples), one plausible approach is to construct ensemble models to utilize the local improvement of each CS model. 


\subsection{Performance of Ensemble Models}

\textbf{Experiments on the Validation Set (12k).} Based on each CS model's performance (e.g., visualized in Figure \ref{fig:ensemble} and \ref{fig:step}) on the validation set, we obtain the weights of the ensemble models using two methods: PCA, and MLP as specified in Section \ref{sec:method}. As shown in Table \ref{table: ind_results}, both of our ensemble models perform 8\%$\sim$10\% better than existing models, the best CS model, the global models, and the simple ensemble approach (majority voting) on the validation set. The improvement is expected as the ensemble weights are constructed based on the performance of CS models on the validation set.

\noindent\textbf{Experiments on the Testing Set (10k).} To show the performance of our ensemble models on random unseen triples, we run the ensemble models on the CC testing set (10k triples with 1,010 random and unique reference images) which has no overlapping with the CC validation set or the CS training set (as stated in Section \ref{sec:dataset}). Table \ref{table: ind_results} shows that both of our ensemble models perform $\sim$5\% better than existing models, the best CS model, the global model, and majority voting. The results show that our analytical ensemble approaches are able to improve global performance (i.e., on random triples), and perform the best on the task of image semantic similarity.

\subsection{Result Analysis and Training Visualization}
\label{sec:train_vis}

\paragraph{Number of Context-Sensitive Models}
Figure \ref{fig:number_proxies} shows the accuracy on the testing set of the three ensemble approaches (Majority Voting, PCA, and MLP) when using different numbers of CS models to form the ensemble models. For a number of selected CS models, we run experiments on all possible combinations. To be specific, when selecting $r$ from the $n=8$ CS models where $r=\{1,2,\ldots,n\}$, we run experiments on all of the $C^n_r$=$\frac{n!}{(n-r)!*r!}$ combinations. For the MLP-based ensemble approach, we repeat the same experiments three times for one given combination, which leads to $3*C^n_r$ runs of experiments for the MLP-based ensemble approach. The results show that the MLP-based approach consistently performs the best, and both of our proposed approaches (MLP-based and PCA-based) perform constantly better than the simple ensemble method, e.g., majority voting. In addition, the results indicate that the accuracy scores on the testing set start to saturate when we use more than six CS models. This might be the reason that the field-sensitive model, forest-sensitive model, and mountain-sensitive model learn similar rules and perform similarly on the testing set. Therefore, assembling these similar CS models might not lead to a significant increase in global accuracy on the testing set. 

\noindent\textbf{Visualization of CS Fine-tuning:} 
We visualize the CS fine-tuning process by showing testing results (on the validation set, 12k random triples), as well as reporting a global accuracy score (on the validation set) at the end of each training epoch. Each scatter point represents the accuracy score of a CS model on all triples $(R_i, x_a, x_b)$ sharing the same reference image $R_i$, and the scatter point is located based on the feature vector of $R_i$ (as stated in Section \ref{sec:method}). As shown in Figure \ref{fig:step}, we highlight the area where we see the local improvement (around $R_i$). Whilst the local performance on the highlighted area has improved, the global accuracy (on the entire validation set) almost remains unchanged, i.e., $\sim$74.5\% and $\sim$76\% for the city- and flower-sensitive model. This shows that CS fine-tuning is able to improve local performance but not global performance, and we show more visualized fine-tuning processes of other CS models in Figure \ref{fig:steps_more} in the appendix.

\subsection{Ablation Studies}

\begin{table}[t]
\centering
\caption{Performance of Ensemble Models (Ranking Block) on Validation Set: 12k triples, and Testing Set: 10k triples.}
\begin{adjustbox}{width=82.5mm}
\begin{tabular}{|c|c|c|c|c|}
\hline
& \multicolumn{1}{|c|}{Best Single} & \multicolumn{1}{c|}{Majority Vote} & \multicolumn{1}{c|}{Ensemble(PCA)} & \multicolumn{1}{c|}{Ensemble(MLP)} \\ \hline
Validation Set & 58.8\% & \multicolumn{1}{c|}{57.1\%} & \multicolumn{1}{c|}{\textbf{84.1\%}} & 78.7\%$\pm$ 0.6\% \\ \hline
Testing Set & 58.7\% & \multicolumn{1}{c|}{57.3\%} & \multicolumn{1}{c|}{69.5\%} & \textbf{71.5\%}$\pm$ 1.0\% \\ \hline
\end{tabular}
\end{adjustbox}
\label{table: ind_results_ranking}
\end{table}

\subsubsection{\textbf{Binary Ranking Blocks}} In addition to embeddings, we also construct ensemble models with the binary classifiers and test the ensembles on the randomly collected triples, i.e., our validation set (12k triples) and testing set (10k triples). As shown in Table \ref{table: ind_results_ranking}, the ensemble model of ranking blocks achieved around 78\%$\sim$84\% on the validation set, and $\sim$70\% on the testing set. The results are significantly better than any of the single CS models (binary classifier) and majority voting. However, the results are 10\%$\sim$15\% worse than using embeddings as shown in Table \ref{table: ind_results}, which is expected as the ranking blocks are trained from scratch using 667 triples only. 
Similarly to Figure \ref{fig:number_proxies} in Section 5.5, we also show the accuracy scores of the ensemble models increase when the number of the CS models (using binary classifiers) increases in Figure \ref{fig:number_proxies_ranking} in the Appendix. Due to the worse performance of ranking blocks on random unseen triples, we focused on the ensemble models constructed using those CS models based on the embedding distances, rather than these binary classifiers.

\subsubsection{\textbf{Cross Validation of CS Models}}
To investigate how the eight selected CS models perform on the other types of unseen reference images, we test each CS model $\mathrm{M}_i$ on all CS datasets $\{D_1, D_2, ..., D_k\}$. As shown in Table \ref{table:cross_val} in the appendix, each CS model performs the best when the reference image is the same as the ones they are trained for, i.e., the accuracy scores on the diagonal are the highest for each CS cluster. Comparing Table \ref{table:cross_val} with Table \ref{table: acc_context_dataset_reviews}, we can make the following observations:
\begin{enumerate}
    \item The fine-tuned CS models performed the best on their corresponding CS data cluster (84\%$\sim$91\%), suggesting some advantages of context-sensitive training (local improvement).
    \item Close examination shows that some CS models perform reasonably well on some other CS data clusters (e.g., the Indoor Model on \#City data cluster), but this does not occur consistently (e.g., the Forest Model on \#Indoor and \#Abstract data clusters). This suggests that (i) our CS models can be used on the data that they have not seen in some cases, and (ii) If we can predict statistically how our CS models will perform on unseen reference images via testing and analysis, we are able to produce a stronger ensemble model.
\end{enumerate}

\subsubsection{\textbf{Impact of Each CS Model}}
To compare how the eight CS models contribute to the ensemble model, we construct ensemble models using seven CS models with one CS model being left out. We run each ensemble model on the testing split of each CS cluster. The results are shown in Table \ref{table:missing_ensemble_on_context_dbs} in the appendix. All of the ensembles perform relatively satisfactorily, even on the left-out and unseen clusters.
One interesting observation is that all ensemble models perform well ($\geq$93\%) on the \#Mountain data cluster, including the ensemble ``No Mountain Model''. This suggests that the knowledge of image similarity in the context of mountains might also be learned from other CS data clusters.
We also test the eight ``Ensemble without $X$ CS model'' on the 10k context-convolute testing set, and the results are shown in Table \ref{table:missing_ind} in the appendix. 
The results show CS models have different impacts on different ensemble strategies, e.g., the mountain-sensitive model is considered the most important for the MLP-based ensemble whilst the PCA-based ensemble might consider the indoor-sensitive model the most important. Compared with the results of the ensemble model using all eight CS models (Table \ref{table: ind_results}), the ensemble models with seven CS models only perform slightly worse on average.

\subsubsection{\textbf{Meta CS models fine-tuned on mixed CS data clusters}}
As inspired by \cite{cvpr_ensemble_proxy} where randomized meta-proxies are shown to be more effective, we run experiments on meta-CS models fine-tuned on meta-CS clusters. In Table \ref{table:meta_class} in the appendix, we show the performance of meta-CS models fine-tuned on two or three CS clusters. To be more specific, we fine-tuned 1) the City/Indoor-sensitive model on the City and Indoor clusters, 2) the Nature-sensitive model on the Mountain, Forest, Ocean, and Field clusters, 3) the Object-sensitive model on the Abstract and Flower clusters. The three meta-CS models achieved 70\%$\sim$73\% on average when applied to other different CS clusters, which was similar to the performance of the directly fine-tuned global models (Table \ref{table:global_model}). Additionally, compared with individual CS models (as shown in Table \ref{table: acc_context_dataset}), the meta-CS models did not perform well on any of the individual CS clusters. The results provided more evidence supporting the observation that fine-tuning a global model on the mixtures of multiple CS clusters could not improve the performance, especially when the number of CS clusters was small. Therefore, when constructing an ensemble model, we used only the CS models, each of which was fine-tuned on only one CS data cluster.

\section{Conclusions}

In this paper, we revisited the problem of image similarity and proposed a solution based on context-sensitive (CS) training datasets that contain image triples $(R, A, B)$ focusing only on a few reference images. We trained a set of CS models, and our tests showed their ability to improve performance locally in their corresponding contexts but not globally when being applied to other contexts. We introduced a new approach to estimate a correctness manifold for each CS model based on imagery features and the testing results of the CS model. The estimated manifolds of CS models enable analytical ensemble strategies that predict the correctness probability of each CS model dynamically for each input triple $(R, A, B)$ and determines the contribution of CS models accordingly. Our extensive experiments showed that our proposed methods performed the best in comparison with all existing models, simple ensemble models, individual CS models, and models directly fine-tuned on random triples (global models).

In addition, we have collected a dataset of 30k labelled triples, facilitating the improvement and comparisons of the task of semantic similarity between images.
In future work, we will further explore the paradigm of constructing ensemble models using CS models, which in many ways bears some similarity to human learning.  

All data, annotations, and source code used for this work can be found in \url{https://github.com/Zukang-Liao/Context-Sensitive-Image-Similarity}.

\newpage
\bibliographystyle{ACM-Reference-Format}
\balance
\bibliography{order}

\newpage
\appendix

\onecolumn
\centering
\section*{Appendix}


\begin{figure}[b]
    \centering
    \includegraphics[height=37mm]{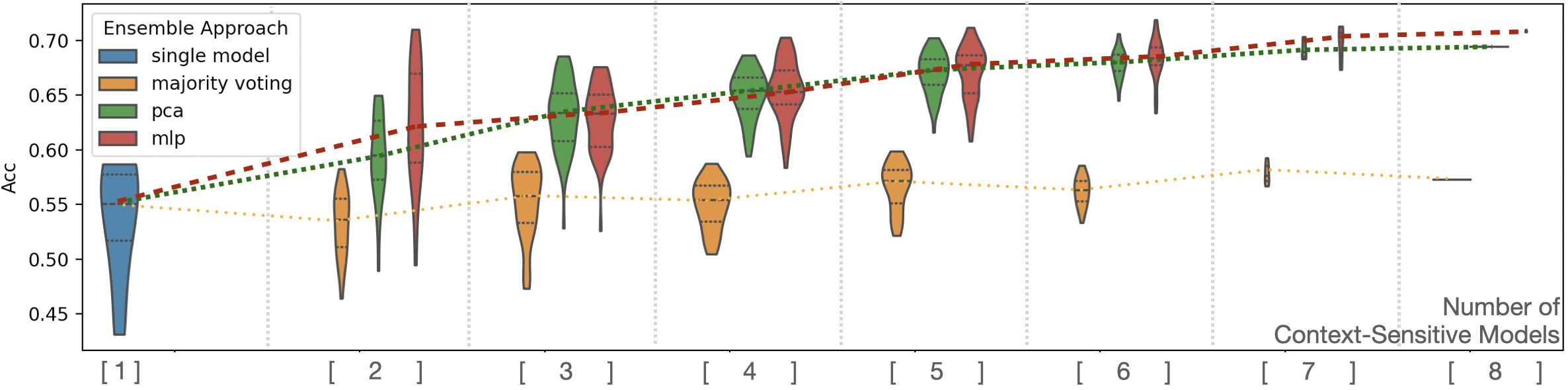}
    \caption{Y-axis: accuracy of the ensemble methods on the testing set (10k). X-axis: the number of CS models used to form the ensemble model. Experiments are run on all the combinations, e.g., when choosing two CS models, we run experiments on all of the $C_{2}^{8}$=$\frac{8!}{2!*6!}$=28 combinations. For MLP, we repeat the same experiment three times for one combination, e.g., for choosing two CS models, we run 3*$C_{2}^{8}$=84 experiments. MLP-based approach consistently performs the best. Dashed lines inside the blobs are the quartiles of the data.}
    \label{fig:number_proxies_ranking}
    \centering
    \includegraphics[height=42mm]{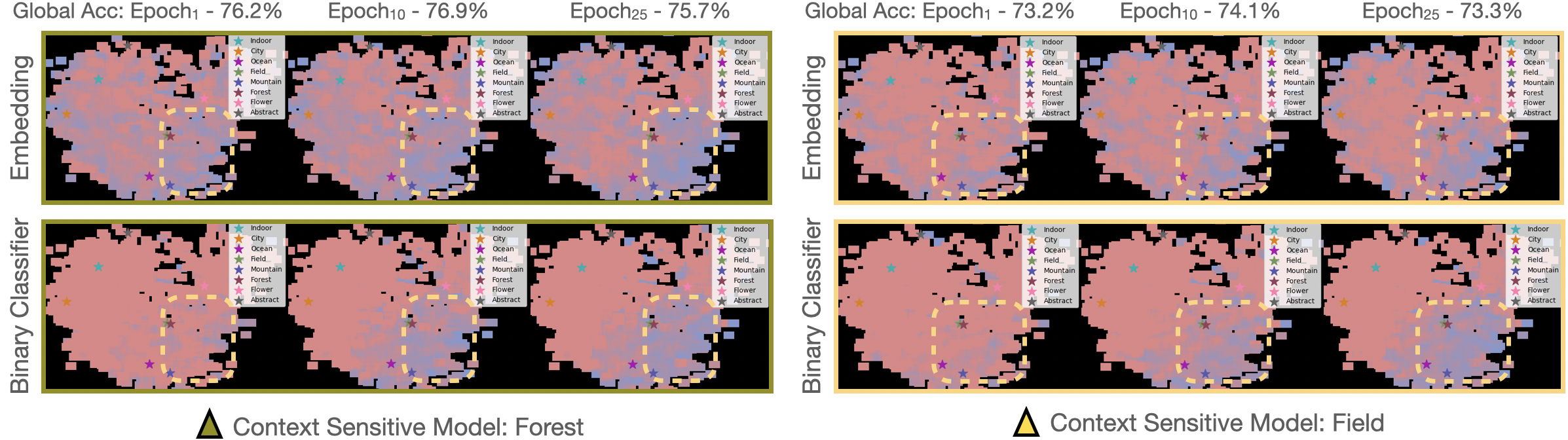}
    \includegraphics[height=42mm]{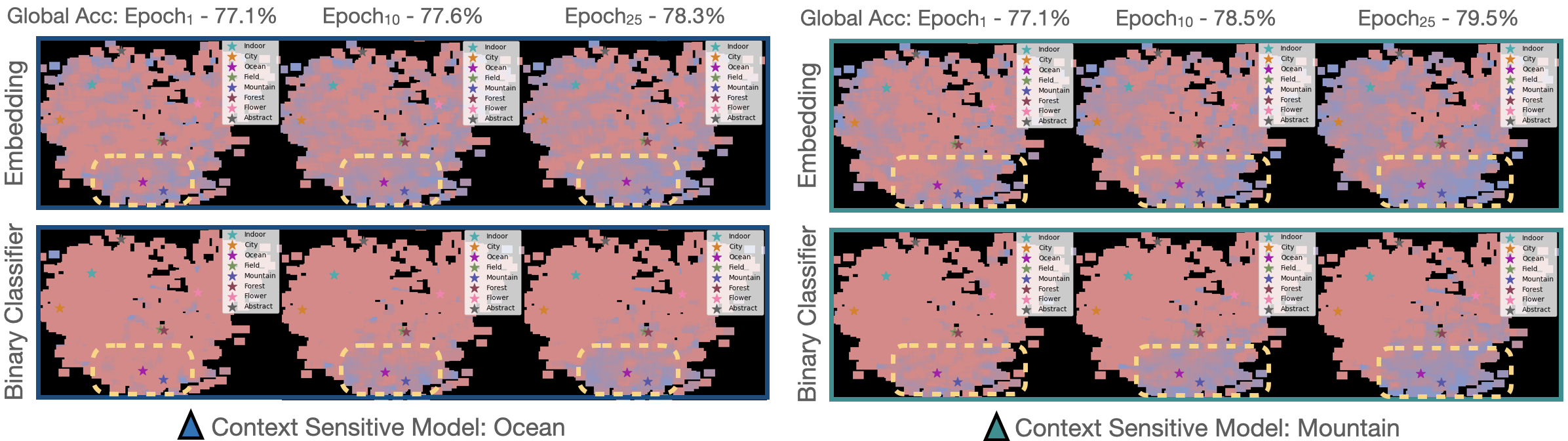}
    \includegraphics[height=42mm]{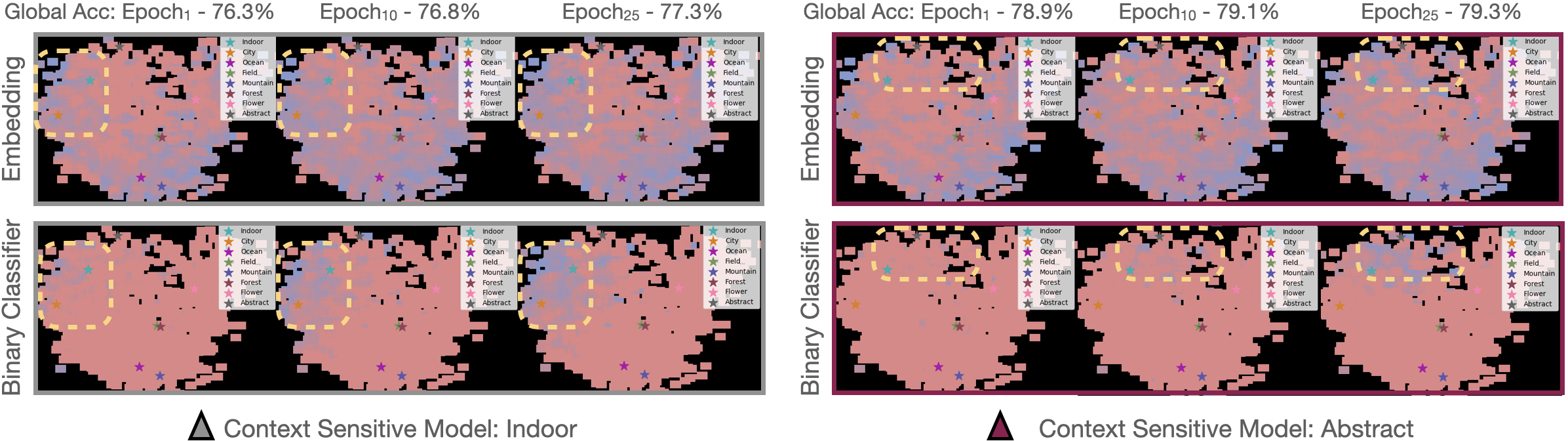}
    \caption{More visualized CS training process: all of the CS models are able to improve local performance (highlighted areas) but not global accuracy. The local improvement of the abstract-sensitive model (on the right) is less noticeable because 1) there are not too many ``abstract" reference images in the validation set, and 2) the ``abstract" images might not be grouped together when applying dimension reduction, e.g., tSNE or PCA.}
    \label{fig:steps_more}
\end{figure}

\begin{table}[h]
\centering
\caption{Cross Validation: performance of each context sensitive models, global model and ensemble models on all of our context datasets.}
\begin{adjustbox}{width=150mm}
\begin{tabular}{|l|c|c|c|c|c|c|c|c|c|}
\hline
\multicolumn{10}{|c|}{\textit{\textbf{Performance of each context-sensitive models on each context dataset}}}\\ \hline
\multicolumn{1}{|c|}{\diagbox[width=36mm]{Model}{Dataset}} & \multicolumn{1}{c|}{\#Indoor}  & \multicolumn{1}{c|}{\#City} & \multicolumn{1}{c|}{\#Ocean}  & \multicolumn{1}{c|}{\#Field} & \multicolumn{1}{c|}{\#Mountain} & \multicolumn{1}{c|}{\#Forest}  & \multicolumn{1}{c|}{\#Flower} & \multicolumn{1}{c|}{\#Abstract}  & \multicolumn{1}{c|}{Average} \\ \hline
\multicolumn{1}{|l|}{Model: Indoor} & \textbf{88.0\%} & 85.7\% & 84.7\% & 71.0\% & 88.9\% & 79.9\% & 74.8\% & 68.8\% & 80.2\% \\ \hline
\multicolumn{1}{|l|}{Model: City} & 55.4\% & \textbf{87.5\%} & 79.0\% & 70.4\% & 82.3\% & 81.7\% & 77.8\% & 69.4\% & 75.4\% \\ \hline
\multicolumn{1}{|l|}{Model: Ocean} & 71.9\% & 82.1\% & \textbf{89.5\%} & 76.7\% & 90.4\% & 65.3\% & 70.0\% & 70.9\% & 77.1\% \\ \hline
\multicolumn{1}{|l|}{Model: Field} & 79.0\% & 77.6\% & 87.1\% & \textbf{90.1\%} & 86.2\% & 81.4\% & 54.7\% & 70.9\% & 78.4\% \\ \hline
\multicolumn{1}{|l|}{Model: Mountain} & 76.0\% & 85.1\% & 85.0\% & 75.5\% & \textbf{90.7\%} & 81.4\% & 82.9\% & 75.4\% & 81.5\% \\ \hline
\multicolumn{1}{|l|}{Model: Forest} & 57.8\% & 76.4\% & 75.4\% & 83.3\% & 87.7\% & \textbf{86.2\%} & 86.2\% & 58.3\% & 76.4\% \\ \hline
\multicolumn{1}{|l|}{Model: Flower} & 59.9\% & 70.4\% & 76.0\% & 76.1\% & 82.6\% & 79.0\% & \textbf{86.8\%} & 52.3\% & 72.9\% \\ \hline
\multicolumn{1}{|l|}{Model: Abstract} & 57.2\% & 77.6\% & 81.7\% & 77.3\% & 90.6\% & 84.7\% & 72.4\% & \textbf{83.8\%} & 78.2\% \\ \hline
\multicolumn{1}{|l|}{Average} & 68.2\% & 80.3\% & 82.3\% & 77.6\% & 87.4\% & 80.0\% & 75.7\% & 68.7\% & - \\ \hline
\multicolumn{1}{|l|}{Average wo diagonal} & 65.3\% & 79.3\% & 81.3\% & 75.7\% & 87.0\% & 79.1\% & 74.1\% & 66.6\% & - \\ \hline
\multicolumn{10}{|c|}{\textit{\textbf{Performance of Our Global, and Ensemble Models}}}\\ \hline
\multicolumn{1}{|l|}{Model: Global} & 57.5\% & 76.4\% & 82.9\% & 74.6\% & 84.1\% & 74.9\% & 64.0\% & 67.3\% & 72.7\% \\ \hline
\multicolumn{1}{|l|}{Ensemble (PCA)} & 76.0\% & 86.9\% & 88.6\% & 85.1\% & 94.0\% & 84.4\% & 86.2\% & 80.5\% & 85.2\% \\ \hline
\multicolumn{1}{|l|}{Ensemble (MLP)} & 79.7\% & 85.1\% & 87.5\% & 89.5\% & 94.0\% & 86.8\% & 86.2\% & 81.2\% & \textbf{86.3\%} \\ \hline
\end{tabular}
\end{adjustbox}
\label{table:cross_val}

\centering
\caption{Ensemble (MLP-based) models with one missing CS model, performance on each context dataset.}
\begin{adjustbox}{width=150mm}
\begin{tabular}{|l|c|c|c|c|c|c|c|c|c|}
\hline
\diagbox[width=36mm]{Ensemble}{Cluster} & \#Indoor & \#City & \#Ocean & \#Field & \#Mountain & \#Forest & \#Flower & \#Abstract & Average \\ \hline
No Indoor Model & 76.0\% & 83.8\% & 86.8\% & 89.2\% & 93.7\% & 86.2\% & 87.1\% & 81.2\% & 85.5\% \\ \hline
No City Model & 81.7\% & 83.5\% & 87.4\% & 88.3\% & 93.4\% & 86.2\% & 87.7\% & 80.8\% & 86.1\% \\ \hline
No Ocean Model & 78.7\% & 85.1\% & 85.6\% & 90.7\% & 93.4\% & 87.7\% & 87.6\% & 81.3\% & 86.3\% \\ \hline
No Field Model & 73.6\% & 84.7\% & 86.2\% & 86.6\% & 93.7\% & 85.0\% & 87.9\% & 80.5\% & 84.8\% \\ \hline
No Mountain Model & 79.3\% & 84.4\% & 87.4\% & 90.1\% & 92.8\% & 86.2\% & 86.2\% & 80.5\% & 85.9\% \\ \hline
No Forest Model & 78.1\% & 86.5\% & 88.6\% & 89.8\% & 93.4\% & 85.9\% & 85.2\% & 80.8\% & 86.0\% \\ \hline
No Flower Model & 80.2\% & 85.6\% & 88.0\% & 89.5\% & 93.7\% & 85.9\% & 85.5\% & 81.7\% & 86.3\% \\ \hline
No Abstract Model & 80.2\% & 85.6\% & 87.4\% & 87.7\% & 93.1\% & 86.2\% & 86.5\% & 78.9\% & 85.7\% \\ \hline
Average & 78.5\% & 84.9\% & 87.2\% & 89.0\% & 93.5\% & 86.2\% & 86.7\% & 80.7\% & - \\ \hline
\end{tabular}
\end{adjustbox}
\label{table:missing_ensemble_on_context_dbs}

\caption{Performance of Ensemble Models with one context-sensitive model being left out on Testing Set (10k)}
\centering
\begin{adjustbox}{width=150mm}
\begin{tabular}{|c|c|c|c|c|c|c|c|c|c|}
\hline
Missing Model: & \cancel{Indoor} & \cancel{City} & \cancel{Ocean} & \cancel{Field} & \cancel{Mountain} & \cancel{Forest} & \cancel{Flower} & \cancel{Abstract} & No Missing \\ \hline
Ensemble PCA & \textbf{81.9\%} & 82.0\% & 82.1\% & 82.4\% & 82.4\% & 82.7\% & 82.6\% & \textbf{82.9\%} & 83.3\% \\ \hline
Ensemble MLP & 83.9\% & 83.8\% & 84.2\% & \textbf{84.5\%} & \textbf{83.7\%} & 84.1\% & 84.2\% & 83.9\% & 84.7\% \\ \hline
\end{tabular}
\end{adjustbox}
\label{table:missing_ind}

\centering
\caption{Performance of meta-CS Models (ResNet18-Place365) trained on combined CS clusters, tested on each CS cluster. Nature: Ocean / Field / Mountain / Forest, Object: Flower / Abstract.}
\begin{adjustbox}{width=150mm}
\begin{tabular}{|l|c|c|c|c|c|c|c|c|c|}
\hline
\diagbox[width=36mm]{CS Model}{CS Cluster} & \#Indoor & \#City & \#Ocean & \#Field & \#Mountain & \#Forest & \#Flower & \#Abstract & Average \\ \hline
City/Indoor & 84.4\% & 82.1\% & 63.2\% & 49.3\% & 80.8\% & 73.7\% & 72.7\% & 56.2\% & 70.3\% \\ \hline
Nature & 78.4\% & 77.0\% & 68.9\% & 59.7\% & 86.8\% & 76.9\% & 63.1\% & 77.2\% & 73.5\% \\ \hline
Object & 60.2\% & 77.9\% & 74.3\% & 67.8\% & 76.6\% & 76.9\% & 68.2\% & 71.5\% & 71.7\% \\ \hline
\end{tabular}
\end{adjustbox}
\label{table:meta_class}

\centering
\caption{Performance of different context-sensitive models (trained on the context training dataset) on the corresponding testing dataset.}
\begin{adjustbox}{width=150mm}
\begin{tabular}{|lccccccccc|}
\hline

\multicolumn{10}{|c|}{\textit{\textbf{Our Context-Sensitive Models - Different Pre-trained Architectures}}}\\ \hline
\multicolumn{1}{|m{32mm}|}{Reference Image} & \multicolumn{1}{c|}{\#Indoor}  & \multicolumn{1}{c|}{\#City} & \multicolumn{1}{c|}{\#Ocean}  & \multicolumn{1}{c|}{\#Field} & \multicolumn{1}{c|}{\#Mountain} & \multicolumn{1}{c|}{\#Forest}  & \multicolumn{1}{c|}{\#Flower} & \multicolumn{1}{c|}{\#Abstract}  & \multicolumn{1}{c|}{Average} \\ \hline
\multicolumn{1}{|l|}{VGG11-ImageNet} & \multicolumn{1}{c|}{86.5\%} & \multicolumn{1}{c|}{84.2\%} & \multicolumn{1}{c|}{85.9\%} & \multicolumn{1}{c|}{88.3\%} & \multicolumn{1}{c|}{86.9\%} & \multicolumn{1}{c|}{76.4\%} & \multicolumn{1}{c|}{67.2\%} & \multicolumn{1}{c|}{76.7\%} & 81.5\% \\ \hline
\multicolumn{1}{|l|}{VGG13-ImageNet} & \multicolumn{1}{c|}{85.9\%} & \multicolumn{1}{c|}{86.9\%} & \multicolumn{1}{c|}{84.5\%} & \multicolumn{1}{c|}{89.0\%} & \multicolumn{1}{c|}{87.1\%} & \multicolumn{1}{c|}{77.5\%} & \multicolumn{1}{c|}{68.3\%} & \multicolumn{1}{c|}{76.9\%} & 82.0\% \\ \hline
\multicolumn{1}{|l|}{ResNet50-ImageNet} & \multicolumn{1}{c|}{41.6\%} & \multicolumn{1}{c|}{66.8\%} & \multicolumn{1}{c|}{58.6\%} & \multicolumn{1}{c|}{74.3\%} & \multicolumn{1}{c|}{80.1\%} & \multicolumn{1}{c|}{60.0\%} & \multicolumn{1}{c|}{62.6\%} & \multicolumn{1}{c|}{49.6\%} & 61.7\% \\ \hline
\multicolumn{1}{|l|}{ResNet34-ImageNet} & \multicolumn{1}{c|}{50.3\%} & \multicolumn{1}{c|}{71.0\%} & \multicolumn{1}{c|}{60.2\%} & \multicolumn{1}{c|}{72.8\%} & \multicolumn{1}{c|}{83.9\%} & \multicolumn{1}{c|}{63.3\%} & \multicolumn{1}{c|}{62.6\%} & \multicolumn{1}{c|}{53.4\%} & 64.7\% \\ \hline
\end{tabular}
\end{adjustbox}
\label{table: acc_context_dataset_more}
\end{table}

\end{document}